\documentclass[lettersize,journal]{IEEEtran}
\usepackage{amsmath,amsfonts}
\usepackage{algorithmic}
\usepackage{algorithm}
\usepackage{array}
\usepackage[caption=false,font=normalsize,labelfont=sf,textfont=sf]{subfig}
\usepackage{textcomp}
\usepackage{stfloats}
\usepackage{url}
\usepackage{verbatim}
\usepackage{graphicx}
\usepackage{cite}
\usepackage{xurl}   
\usepackage{hyperref}

\usepackage{booktabs}
\usepackage{tabularx}
\usepackage{makecell}
\usepackage{multirow}

\begin{document}

\title{Context-Aware Intelligent Vehicles}

\author{
\IEEEauthorblockN{
Liangkai~Liu\IEEEauthorrefmark{1},
Shuyao~Shi\IEEEauthorrefmark{1},
Mingke~Wang\IEEEauthorrefmark{1},
Noah~T.~Curran\IEEEauthorrefmark{1},
Chuan~Li\IEEEauthorrefmark{2}, 
Fan~Bai\IEEEauthorrefmark{2}, and
Kang~G.~Shin\IEEEauthorrefmark{1}
}
\IEEEauthorblockA{
\IEEEauthorrefmark{1}Department of Computer Science and Engineering, University of Michigan, USA
\\
\IEEEauthorrefmark{2}General Motors, Warren, Michigan, USA
}
}




\maketitle

\begin{abstract}
Intelligent vehicles increasingly support adaptive applications beyond driving themselves, ranging from context-aware ADAS and automated driving to in-cabin monitoring and fleet management, all under tight requirements on accuracy, latency, cost, and reliability. Meeting these requirements is challenging because vehicles operate in complex, uncertain, and rapidly changing environments while running on resource-constrained computing platforms. This paper argues that \textit{context}—situational factors that give meaning to sensor signals and constrain decisions—should be treated as a first-class principle for next-generation vehicle systems, and operationalized as a unified, shared state for learning, risk assessment, and closed-loop control across the software stack. We systematically review state-of-the-art (SOTA) context-aware methods spanning (i) environment understanding, (ii) planning and control, (iii) safety and security, and (iv) connected vehicles. Based on a trend analysis of context-aware design, we identify four key technical challenges in building a general contextual engine for future intelligent vehicles: multimodal context fusion, temporal context modeling, handling rare events, and collaborative context sharing. We hope this survey will motivate the development of robust and efficient context-aware vehicle applications.

\end{abstract}


\section{Introduction}

Advances in sensing, computer vision, machine learning (including deep 
neural networks and foundation models), hardware acceleration, and vehicle 
connectivity (e.g., DSRC, C-V2X, 5G) have propelled intelligent and 
autonomous vehicles to the forefront of research and industry
\cite{yurtsever2020survey, janai2020computer, geiger2012we, liu2020computing, kato2015open, abboud2016interworking, hwang2024emma, bar2025navigation}.

As autonomy and connectivity mature, vehicles are evolving from transportation 
tools into context-driven intelligent platforms that enable novel applications 
spanning safety, comfort, and mobility services, while operating under strict 
constraints on accuracy, latency, cost, and reliability. On the safety side, 
context-aware ADAS and autonomous driving increasingly fuse HD maps, onboard 
perception, environment/road-condition sensing, and driver-state signals to 
estimate risk and adapt warning/intervention thresholds, improving robustness to 
diverse operating conditions and edge cases, such as vulnerable road users (VRUs) 
and work zones
\cite{kalantari2025much,wang2015forward,habibi2018context,NHTSA2024NCAP_ADAS_Roadmap,ViaSight2025WorkZoneADAS,Bhandare2025ContextAwareADAS}.
Beyond safety, in-vehicle infotainment and human–machine interfaces (HMI) 
are becoming environmental-aware, skipping low-priority interactions during demanding 
maneuvers and leveraging navigation/V2X communications to provide lightweight, 
situational cues (e.g., school zones and adverse weather) without increasing 
distraction~\cite{Sholichin2023HMIHUN,macario2009vehicle,Subiksha2025HMIPrinciples,fernandez2019contextual,CarSifu2021AudiCV2XSchoolSafety}. 
Finally, with driver safeguards, third-party services can use the driving context for 
wellness monitoring and emergency response (e.g., automated crash notification), 
context-aware travel assistance, and fleet operations such as risk-aware routing 
and fatigue-aware planning—highlighting the need for strong privacy, security, 
and governance when handling sensitive logs and incident data~\cite{NEC2017SelfMonitoringServices,EC2025eCallInteroperableEUWide,Smith2019EVRoadTrip,Telematica2024DriverHealthWellness,Plante2024}.

A key barrier to meeting these requirements is the complexity of the 
physical world: vehicles must reason about heterogeneous sensors, uncertain 
dynamics, diverse traffic participants, changing weather/lighting, and 
evolving regulations while running on space/cost/resource-constrained 
platforms~\cite{yurtsever2020survey,liu2020computing}. 
As a result, understanding ``contexts''--the ability to capture 
and exploit the situational factors that shape perception, prediction, 
planning, and control--has become a first-class design principle for 
next-generation vehicle systems.

We define {\em context} as the set of relevant interior and exterior conditions 
and factors that give meaning to, influence, or constrain an action, 
decision, or interpretation. Context is inherently situational 
(tied to time, place, and conditions) and can be explicit (e.g., temperature, 
location, time) or implicit (e.g., social norms, hidden states). 
More importantly, context changes interpretation: the same signal can 
imply different actions under different conditions (e.g., a pedestrian near
a curb at noon vs.~at night in heavy rain). In this paper, we 
represent context as a machine-usable state that conditions 
learning, risk assessment, and control across the autonomy stack.

\begin{figure*}[!htp]
	\centering
	\includegraphics[width=\textwidth]{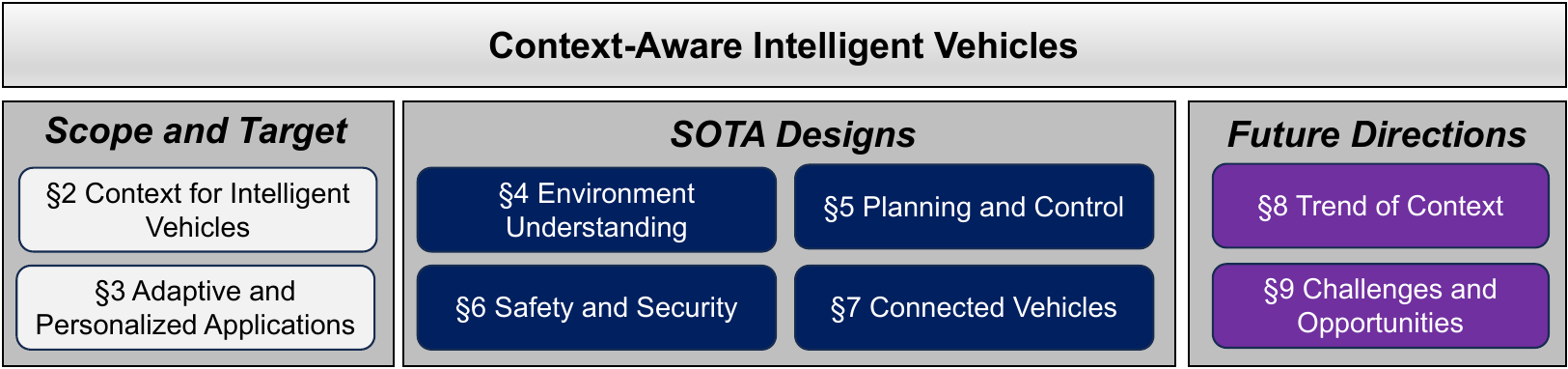}
	\caption{Structure of the context-aware intelligent vehicles.}
	\label{fig:paper-structure}
\end{figure*}

In this paper, we present a comprehensive review on state-of-the-art (SOTA) context-aware methods.
The idea of context-aware designs has already been considered in many applications and systems supporting AI-enabled autonomy~\cite{TITS22Context3D, HydraFusion}. In environmental sensing and perception, context signals like weather, road geometry, lanes, and driver behavior can be used to determine the fusion of multi-modality sensing-based detection and prediction~\cite{HydraFusion, TITS22Context3D, TrafficAug, MMTL-UniAD}. For planning and control of intelligent vehicles, context provides complementary information on dynamic conditions and user preference over traditional kinematic and rule-based methods~\cite{ped-speed-2022, llm-social-2025,caps-2025,predictive-planning-2024,meta-rl-2019, energy-aware-2023}. For instance, context like weather and road conditions could help to generate a trajectory which is safer and more efficient~\cite{robust-nav-2025}. Moreover, it enhances personalized applications like parking to become more convenient for people with special needs~\cite{park4u-2021}. Context also enhances the vehicle's safety and security~\cite{chen2025insight,tian2025context}. By integrating context signals like semantics, text descriptions, weather, traffic, and road inclination into the risk assessment and anomaly detection, collisions can be reduced, and anomalous data groups can be accurately identified~\cite{tian2025context,chen2022context,chen2024context}. Beyond the ego vehicle, contextual information has become increasingly important for connected vehicles~\cite{cress2023intelligent, shi2024soar,zhou2020evolutionary}. Context-awareness is used in connected vehicles through an end-to-end pipeline that defines how information is acquired, communicated, fused, and ultimately used by driving functions~\cite{hu2022where2comm,wang2023vimi,chaudhry2023toward,li2024di,hu2024communication,zhang2023robust}.

By analyzing the trend of context, we identify open research challenges and opportunities for the development of a comprehensive contextual engine for intelligent vehicles.
We observe several trends in the modeling and usage of context. 
Intelligent vehicle stacks are increasingly built around context-rich modeling, moving beyond frame-by-frame perception and fixed heuristics toward temporally grounded, risk-aware, and adaptive autonomy~\cite{robust-nav-2025,tian2025context}. Learned scene/agent representations replace many hand-tuned rules, while temporal fusion and world models lift perception from 3D snapshots to 4D state over time for more stable tracking and intent prediction~\cite{bar2025navigation, MMTL-UniAD}. Context also drives efficiency, with ROI-aware and anytime inference focusing compute and sensing on safety-critical regions to reduce latency and bandwidth~\cite{energy-aware-2023, Bhandare2025ContextAwareADAS}. In parallel, safety shifts from static thresholds to ODD- and uncertainty-aware behavior, and robustness improves via cross-sensor and physics/temporal consistency checks~\cite{HydraFusion,zhang2023robust}. Overall, context links reliability, real-time efficiency, and clearer safety rationale under operational constraints.

The goal of the contextual engine is to maintain a coherent, continuously updated understanding of the external scene (infrastructure, road semantics, dynamic agents) and internal state (ego motion, system health, and potentially driver condition) by integrating signals across perception, prediction, planning/control, and safety/security.
We identified four research challenges and opportunities: (i) multimodal fusion must reconcile heterogeneous sensors and priors with differing formats, latencies, and failure modes to produce a consistent context representation; (ii) temporal modeling must capture multi-horizon dynamics and uncertainty to support intent and trajectory forecasting beyond static snapshots; (iii) rare-event handling requires resource-aware operation that runs efficiently in routine driving yet rapidly escalates sensing and inference under elevated risk, while degrading gracefully under missing or unreliable context; and (iv) collaborative context sharing promises broader situational awareness through V2X, but demands common abstractions, bandwidth/latency-aware protocols, and strong trust and conflict-resolution mechanisms. Together, these challenges define the pathway to more reliable, anticipatory, and efficient vehicle intelligence.

The paper is organized as shown in Figure~\ref{fig:paper-structure}.
Section~\ref{sec:context} and Section~\ref{sec:apps} cover the scope and target of this paper by defining context and discussing the novel adaptive and personalized applications that can be enabled by context.
Next, we cover the context-aware design in the state-of-the-art (SOTA) intelligent vehicle pipeline.
We present the use of context in environmental sensing and perception~\ref{sec:perception}, planning and control~\ref{sec:planning-control}, safety and security~\ref{sec:safety}, and connected vehicles~\ref{sec:v2x}.
Next, we summarize emerging trends of context-aware design in Section~\ref{sec:trend}, and present challenges and research opportunities for building a contextual engine for intelligent vehicles in Section~\ref{sec:challenge}. The paper concludes with Section~\ref{sec:conclusion}.


\section{\emph{Context} for Intelligent Vehicles}
\label{sec:context}

\vspace{2mm}
\noindent \textbf{Intelligent vehicles} are emerging as a core computing and sensing platform in modern commercial vehicles, spanning the full spectrum of driving automation from SAE Level~0 (human-driven) to Level~5 (fully automated)~\cite{on2021taxonomy, gusikhin2008intelligent}. In this review, we define an intelligent vehicle as a vehicle that exhibits three key characteristics:
(i) \textit{Automation capability}: supports functions across Levels~0--5, ranging from driver assistance to full autonomy;
(ii) \textit{Multi-source data fusion}: ingests and integrates heterogeneous data from onboard (ego) sensors, infrastructure and V2X inputs, and external Internet/cloud services (e.g., open APIs);
(iii) \textit{Personalized services}: enables novel, context-aware applications that adapt to individual users, preferences, and operational needs through customization and personalization.

\vspace{2mm}
\noindent \textbf{Context} is the set of relevant surrounding conditions and factors that give meaning to, influence, or constrain an action, decision, or interpretation. It is inherently \emph{situational}—bound to a particular time, place, and set of conditions.
Context can be \emph{explicit} (measurable variables such as temperature, location, or time) or \emph{implicit} (social norms, hidden states, or conventions). Crucially, context \emph{changes interpretation}: the same signal can imply different actions under different conditions (e.g., a pedestrian standing near a curb at noon versus at night in heavy rain). 


\begin{figure*}[!htp]
	\centering
	\includegraphics[width=.8\textwidth]{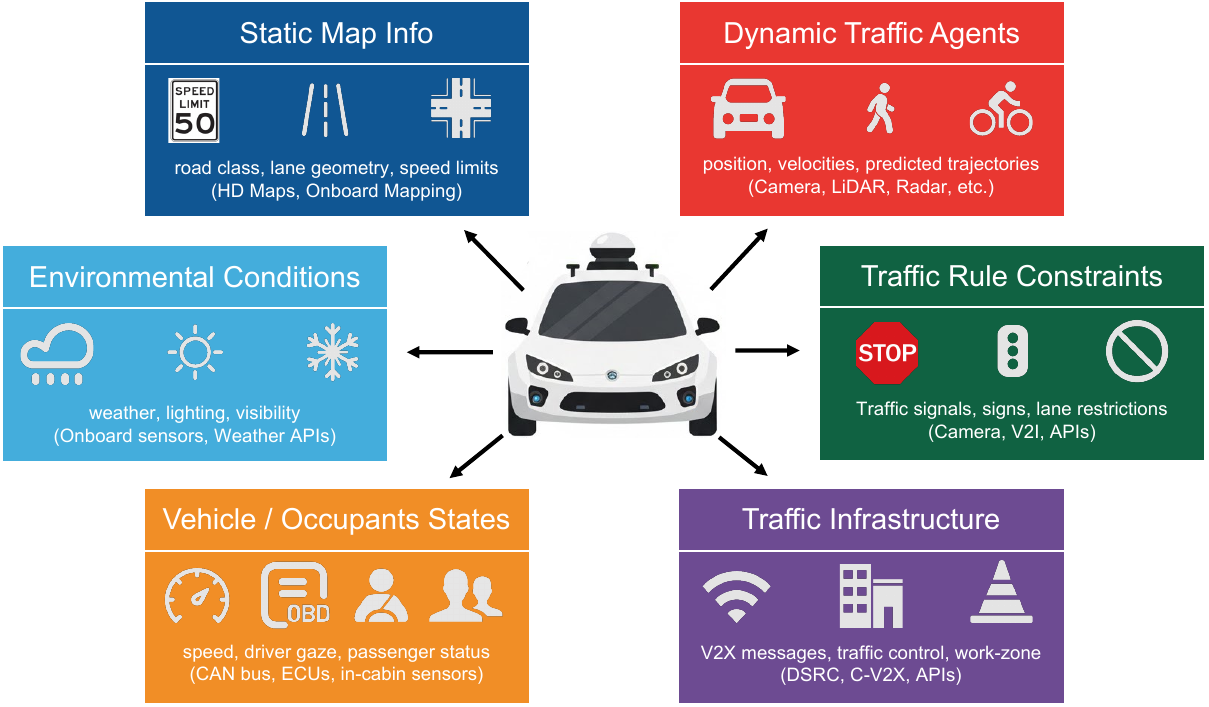}
	\caption{Context map for intelligent vehicles.}
	\label{fig:context-map}
\end{figure*}

\vspace{2mm}
\noindent \textbf{Driving Context State \(\mathbf{C}(t)\).} 
Figure~\ref{fig:context-map} and Table~\ref{tab:context-definition} formalize the driving context as a time-varying state vector \(\mathbf{C}(t)\) that concatenates six complementary dimensions, each populated by variables with different sampling rates and uncertainty profiles. \emph{Static Map Info} provides priors—road class, lane geometry, intersection topology, and speed limits—typically sourced from HD maps or onboard mapping; these priors anchor localization and constrain feasible maneuvers. \emph{Dynamic Agents} capture nearby vehicles and vulnerable road users via perception outputs (positions/velocities and predicted trajectories) from cameras, LiDAR, and radar; these estimates evolve fastest and dominate short-horizon risk. \emph{Environmental Conditions} encode weather, illumination, visibility, and surface state, combining onboard environment sensors with external weather APIs to modulate sensing fidelity and traction-aware planning. Traffic Rule Constraints comprise machine-readable traffic signals, signs, lane restrictions, and right-of-way rules detected by cameras and reinforced via V2I, gating maneuver legality. \emph{Ego State} aggregates the vehicle’s kinematics (speed, yaw rate, acceleration), localization quality, system health, and compute load from CAN/ECUs and the localization stack, exposing self-awareness for timing and safety envelopes. Finally, Traffic Infrastructure Data brings V2X messages, traffic-control feeds, and work-zone beacons (DSRC/C-V2X/infrastructure APIs) that extend situational awareness beyond line-of-sight. In practice, \(\mathbf{C}(t)\) results from time alignment and quality-weighted fusion of these sources and is consumed by perception–prediction–planning–monitoring modules to improve robustness, legality, and real-time performance. For example, with rich contextual information about road conditions (hazards, elevation changes, sharp turns, etc.) and occupants (e.g., a pregnant passenger or an infant), an intelligent vehicle can recommend navigation routes and parking plans that are more convenient and comfortable.

\begin{table*}[t]
\centering
\small
\caption{A structured view of driving context as a state vector $\mathbf{C}(t)$.}
\label{tab:context-definition}
\begin{tabular}{p{0.20\linewidth} p{0.42\linewidth} p{0.28\linewidth}}
\toprule
\textbf{Dimension} & \textbf{Example Variables} & \textbf{Typical Sensors / Sources} \\
\midrule
Static Map Info &
road class, lane geometry, intersection topology, speed limits &
HD maps, onboard mapping \\
\addlinespace[0.35em]
Dynamic Agents &
positions, velocities, predicted trajectories of vehicles/pedestrians &
cameras, LiDAR, radar \\
\addlinespace[0.35em]
Environmental Conditions &
weather, lighting, visibility, road surface &
onboard environment sensors, weather APIs \\
\addlinespace[0.35em]
Traffic Rule Constraints &
traffic signals, signs, lane restrictions, right-of-way rules &
cameras, V2I \\
\addlinespace[0.35em]
Vehicle / Occupants State &
speed, yaw rate, acceleration, localization quality, driver gaze, passenger status, system health, compute load &
CAN bus, ECUs, in-cabin states \\
\addlinespace[0.35em]
Traffic Infrastructure Data &
V2X messages, traffic control center feeds, work-zone beacons &
DSRC, C-V2X, infrastructure APIs \\
\bottomrule
\end{tabular}
\end{table*}

\noindent
This decomposition clarifies both \emph{what} constitutes context and \emph{how} it is acquired. By maintaining $\mathbf{C}(t)$ online, the intelligent vehicle can condition perception (e.g., BEV fusion with map priors), calibrate prediction and risk (e.g., uncertainty under adverse weather), and adapt control policies to the current Operational Design Domain (e.g., school-zone speed limits, lane closures), yielding decisions that are not only accurate but also interpretable and compliant with constraints.

\section{Adaptive and Personalized Applications}
\label{sec:apps}


\subsection{Advanced Driver-Assistance Systems (ADAS) Adaptation}

Modern advanced driver-assistance systems (ADAS) are increasingly adaptive and 
personalized, tuning their behavior to contextual factors and individual drivers. 
Safety-critical functions such as forward collision warning (FCW) and automatic 
emergency braking (AEB) adjust sensitivity thresholds using inputs like road type, 
traffic density, and weather~\cite{kalantari2025much}. For example, a car may warn 
earlier and brake more conservatively on a rainy street than on a dry street. 
Following distance safety is also context-dependent: adaptive cruise systems can 
extend headway at high speeds or in poor visibility, as shorter gaps ($<$2s) 
greatly raise collision risk~\cite{wang2015forward}.

Adaptive ADAS capabilities extend beyond collision avoidance. 
AEB systems increasingly detect vulnerable road users (VRUs) such as pedestrians and 
cyclists~\cite{NHTSA2024NCAP_ADAS_Roadmap}. 
Context-aware modules can also anticipate special hazards, e.g., detecting red-light 
runners or handling temporary work zones where unclear signs or traffic cones challenge lane-keeping automation~\cite{ViaSight2025WorkZoneADAS}. 
In low-friction conditions, vehicles estimate road slipperiness via wheel slip 
sensors and adjust braking or following-distance recommendations accordingly. 
Personalization also includes driver conditions: if sensors detect drowsiness or 
incapacitation, systems such as Emergency Assist autonomously slow down the ego vehicle, maintain lane position, and stop safely~\cite{DouradoRS7Safety}.

To enable these behaviors, ADAS fuses data from multiple sources: high-definition maps 
for road geometry and limits, onboard vision and radar for surrounding vehicles and 
pedestrians, and sensors for weather and road friction~\cite{habibi2018context}.
Driver-monitoring cameras assess attention and alertness, allowing the system to 
adapt alert timing or intervention thresholds~\cite{Bhandare2025ContextAwareADAS}. 
Real-time fusion of these signals supports dynamic risk estimation and 
intent prediction.

The outputs of adaptive ADAS are conveyed either internally or to the driver. 
Vehicles may compute a continuous risk index to trigger context-sensitive warnings 
or mode switches, adjust adaptive cruise headway, or issue alerts such as 
“Road may be icy.” Some research prototypes visualize this feedback as a “risk 
meter”~\cite{ren2025lane}. By integrating environmental, behavioral, and 
driver-state data, adaptive ADAS aims to balance safety, comfort, and user 
trust through personalized, context-aware assistance.



\subsection{Infotainment and In-Vehicle Personalization}

Beyond safety, automotive technology now support personalized service and 
human–machine interfaces (HMI) using in-vehicle edge computing and cloud services~\cite{zhang2018openvdap}. 
In-car infotainment (IVI) adapts to driver workload and preferences to enhance 
convenience without adding distraction. A prime example is a 
\textit{distraction-aware UI}: when cognitive load is high (e.g., heavy traffic or 
complex maneuvers), the system hides non-urgent notifications. 
Modern platforms like Android Automotive implement contextual notification management; 
the Heads-Up Notification (HUN) framework shows the driver only the most important alerts, using the car’s current status and what the driver is doing to decide what matters. It hides minor, non-urgent pop-ups while the vehicle is moving~\cite{Sholichin2023HMIHUN}. This minimizes visual and cognitive load while 
keeping the driver informed, aligning with guidance that HMI feedback should be 
context-aware and timely~\cite{macario2009vehicle,Subiksha2025HMIPrinciples}. 
For instance, the vehicle may automatically enter “do not disturb” in dense traffic
or reduce re-routing prompts during heavy rain, focusing attention on critical 
safety alerts.

IVI also incorporates real-time driver state. If monitoring indicates drowsiness or 
distraction, the interface can simplify and prompt a rest. 
Various types of cars already issue “take a break” suggestions based on fatigue 
indicators. Studies emphasize that once drowsiness is detected, prompt 
countermeasures are crucial, ranging from auditory/tactile warnings to supportive 
automation~\cite{ayas2024drowsiness}. An adaptive IVI might temporarily tighten 
assistance features and lower cruising speed, adjust HVAC to increase alertness, 
or, under high stress, reduce non-essential notifications and play calming 
audio~\cite{ayas2024drowsiness}.

Personalization extends to \textit{context-aware navigation and 
assistance}~\cite{fernandez2019contextual}. GPS and maps can tag cues 
such as school zones, hospitals, or hazardous areas, and adjust HMI policy accordingly. 
Early deployments show C-V2X (cellular vehicle-to-everything) warnings for active 
school zones or stopped school buses; roadside units broadcast signals that trigger 
dashboard and audible alerts urging drivers to slow down~\cite{CarSifu2021AudiCV2XSchoolSafety, liu2016implementation}. 
Weather context can similarly refine guidance: if heavy rain is detected, navigation
may propose safer routes, annotate the map, and reduce spoken directions. 
Other contextual hints include avoiding stadium traffic after events or reminding 
drivers of work-zone speed limits—using lightweight, well-timed cues to support 
attention.

Cabin comfort and driver profiles are increasingly tailored. Vehicles store driver-specific 
preferences for seating, mirrors, climate, audio, and displays, restoring them when 
a driver is recognized (key fob, phone, or face). For example, Driver 
Personalization System loads seating, steering wheel, temperature, and cluster 
settings for each user~\cite{Mazda2025CX70DriverPersonalization}.
Occupant sensing can identify who sits where and apply the corresponding comfort 
profiles, which can vary by time of day (e.g., morning “news + cool” vs.~evening
“relax + warm”).

In summary, infotainment and comfort features are becoming deeply personalized, 
using driver state, passenger identity, and situational context (location, time, 
weather) to adapt the UI, notifications, climate, and routing. 
The goal is to optimize both safety and comfort, delivering an in-cabin experience 
that responds moment by moment to occupants’ needs.


\subsection{Third-Party Applications Leveraging Driving Context Data}

With driver consent and proper safeguards, third-party services can use the vehicle
context and driving data to deliver adaptive, personalized benefits.

\vspace{2mm}
\noindent\textbf{Wellness and Safety Monitoring Services:} Context data can trigger 
protective actions, such as notifying familty or relatives when extreme drowsiness or erratic 
driving is detected~\cite{NEC2017SelfMonitoringServices}. eCall, standard on new EU 
cars, automatically contact 112 after severe crashes, sending location and crash data 
to speed response and reduce fatalities~\cite{EC2025eCallInteroperableEUWide}. 
Third-party services can additionally alert designated contacts. These interventions leverage driver-state and crash signals for timely assistance.

\vspace{2mm}
\noindent\textbf{Fleet Operations and Logistics:} Fleets use driving context to 
optimize safety and efficiency via risk-aware routing (traffic, weather, crime), 
proactive rerouting for congestion or ice, and fatigue-aware rest-stop 
recommendations~\cite{Telematica2024DriverHealthWellness}. Telematics dashboards 
support coaching (hard braking, speeding, seatbelt use) with real-time in-cab feedback, 
improving habits and reducing incidents. Condition-based maintenance schedules service 
by usage and stress (e.g., brake wear from hilly urban routes), minimizing downtime.

\vspace{2mm}
\noindent\textbf{Personalized Commerce and Travel Services:} Connected-car data powers 
context-aware stops and recommendations. EV trip planners can pair charging with rest 
when fatigue is likely, aligning with guidance to break every 2-3 
hours~\cite{Smith2019EVRoadTrip}. Apps may suggest safer parking spot or 
timely coffee orders along routine routes, which will be hidden to avoid intrusiveness while merging location, timing, and driver state.

\vspace{2mm}
\noindent\textbf{Accident Forensics and Law Enforcement Assistance:} With consent or 
legal authority, telematics can reconstruct pre-crash events (speed, steering, braking) 
to clarify fault and support court cases~\cite{CarusoLaw2024TelematicsPI}. 
Future systems could securely share recent driving snapshots during traffic stops, 
though privacy and legal safeguards are essential. Modern vehicles function as data 
recorders; when used responsibly, their logs can improve accountability and road safety.

Overall, adaptive and personalized applications --- whether onboard in the vehicle or 
provided by third parties --- represent a significant evolution in driving. 
They harness a wealth of sensor data and contextual information to tailor 
the driving experience in real time, enhancing safety, comfort, and 
efficiency~\cite{garzon2012intelligent}. From smart ADAS that adjusts to weather and 
driver state, to infotainment that knows when to keep quiet, and services that reward 
safe habits or assist in emergencies, personalization is making transportation more 
responsive to human needs. This human-centric approach, grounded in context-awareness, 
is a hallmark of next-generation automotive systems~\cite{Plante2024}. As vehicles 
continue to become more connected and intelligent, we can expect these adaptive 
applications to grow even more capable, ultimately converging towards a safer and 
more personalized mobility ecosystem.

\section{Context-Aware Environment Sensing and Understanding}
\label{sec:perception}

\begin{table*}[t]
\centering
\small
\caption{Representative context-aware perception.}
\label{tab:context-methods}
\begin{tabularx}{\linewidth}{@{}p{0.18\linewidth} p{0.26\linewidth} p{0.34\linewidth} p{0.18\linewidth}@{}}
\toprule
\textbf{Work} & \textbf{Key Context Signals} & \textbf{Mechanism} & \textbf{Impact} \\
\midrule
\textit{HydraFusion~\cite{HydraFusion}} &
Weather, lighting, scene type &
Selective fusion that switches among early/late/hybrid fusion conditioned on context &
$\uparrow$ detection robustness in adverse conditions without extra compute \\
\addlinespace[0.35em]
\textit{Context-3D~\cite{TITS22Context3D}} &
Road geometry, object relations &
Combines contextual features to inform depth/scale priors for 3D reasoning &
$\uparrow$ 3D precision for distant/occluded objects \\
\addlinespace[0.35em]
\textit{Traffic-Context~\cite{TrafficAug}} &
Lanes, freespace, traffic participants &
Uses traffic context as constraints when augmenting scenes to keep semantics/plausibility &
$\uparrow$ rare-class recall \\
\addlinespace[0.35em]
\textit{MMTL-UniAD~\cite{MMTL-UniAD}} &
Traffic, driver behavior/emotion, vehicle behavior &
Multi-axis region-attention network to extract global context-sensitive features for multi-task heads &
Improves all four tasks \\
\bottomrule
\end{tabularx}
\end{table*}

Perception benefits from \emph{context}—signals beyond raw sensor data that 
constrain interpretation, improve robustness, and calibrate confidence. 
Below, we group the literature into four compact categories that cover 
multimodal learning, scene/social priors, adaptive operations, and reliability.

\vspace{2mm}
\noindent\textbf{Unified multimodal \& multitask context.}
Jointly modeling in-cabin and external cues, as well as multiple tasks, 
yields a shared structure that reduces negative transfer and improves data efficiency. 
Frameworks such as \textit{MMTL-UniAD} couple driver state (behavior/emotion) with 
traffic and vehicle context via region attention and dual-branch embeddings for 
balanced shared/specific features, while datasets like \textit{AIDE} provide 
synchronized multi-view, multi-modal annotations to train such holistic models 
\cite{MMTL-UniAD, AIDE}. Language-grounded approaches (e.g., \textit{ContextVLM}) 
further enable zero-/few-shot context recognition with VLMs, lowering supervision 
costs \cite{ContextVLM}.

\vspace{2mm}
\noindent\textbf{Scene, geo, and social context for geometry \& motion.}
Map layout, lanes, and right-of-way rules regularize geometry and depth/scale, 
improving single-image 3D detection and long-range/occluded cases 
\cite{TITS22Context3D}. Forecasting models that encode agent–agent and 
agent–scene interactions (e.g., \textit{PRISC-Net}, \textit{CAPHA}, \textit{CASTNet}) 
generate kinematically feasible, rule-compliant, and context-adaptive trajectories 
\cite{PRISCNet, CAPHA, CASTNet}. Beyond a single region, geo-context studies show 
distribution shifts across locales and motivate geo-aware training/augmentation 
to sustain robustness under deployment shifts \cite{GeoContext}.

\vspace{2mm}
\noindent\textbf{Context-driven fusion.}
Operational policies can be conditioned on context to boost reliability without 
extra computation. \textit{HydraFusion} selects early/late/hybrid sensor fusion 
by weather/lighting, outperforming fixed fusion under adverse conditions 
\cite{HydraFusion}. Traffic-aware augmentation inserts rare objects only in 
semantically plausible locations (lanes/freespace/flows) to lift long-tail recall 
\cite{TrafficAug}. Context-aware data engines (e.g., \textit{ADAM}) use LLMs 
to produce high-level, scene-consistent annotations that accelerate dataset
curation \cite{ADAM}.

\vspace{2mm}
\noindent\textbf{Reliability: uncertainty \& system-level awareness.}
Context should also modulate trust. Uncertainty frameworks adjust confidence with 
visibility, occlusion, and weather to curb overconfidence in degraded sensing 
\cite{PerceptualUncertainty2022}. At the autonomy-stack level, persistent context 
(sensor health, landmark availability) enables behavior-tree or policy switching 
among localization strategies, improving overall resilience; drivable-area estimation 
(e.g., \textit{DriveSpace}) similarly stabilizes freespace in complex layouts 
\cite{BehaviorTreeLocalization2021, DriveSpace2019}. Multi-task scene understanding 
further shares context to recognize scenarios like construction zones while 
detecting relevant entities \cite{ContextAwareMTL2020}.

\vspace{2mm}
\noindent\textbf{Takeaway.}
Fewer, broader principles emerge: (i) learn shared structure across modalities/tasks; 
(ii) encode scene/geo/social priors to resolve ambiguity and predict motion;
(iii) route fusion and data workflows by context; and (iv) calibrate uncertainty 
and switch system behaviors using operational context. Together these shift 
perception from frame-centric to world-model-centric autonomy.


\section{Context-Aware Planning and Control}
\label{sec:planning-control}
The planning and control system bridges perception and actuation. Traditional approaches rely on kinematic models and rule-based methods that struggle with the variability of real-world traffic. Context-aware techniques address this gap by adapting to dynamic conditions and user preferences.
We organize recent advances into four dimensions: \textit{social and semantic context}, \textit{adaptive decision-making}, \textit{personalization}, and \textit{learning-based adaptation}.

\vspace{2mm}
\noindent\textbf{Social and Semantic Context Integration.}
Geometric planners ignore the implicit social contracts that govern human driving.
LLM-based approaches~\cite{llm-social-2025} translate high-level cues---``aggressive driver behind,'' ``school zone''---into behavioral adjustments such as larger following distances or reduced lane-change aggressiveness.
The \textit{Pedestrian-Aware Speed Controller}~\cite{ped-speed-2022} stacks three speed limits: legal, density-based (fewer pedestrians permit higher speeds), and proximity-based (slow further when individuals are near). The goal is \textit{perceived safety}: pedestrians should feel safe, not just be safe.

\vspace{2mm}
\noindent\textbf{Adaptive Decision-Making Under Uncertainty.}
\textit{Predictive Motion Planning}~\cite{predictive-planning-2024} generates predictions of surrounding vehicles' movement using multi-modal data via an Interaction-Aware IMM Kalman Filter, then optimizes the trajectories with MPC under tunable safety margins. The framework handles both maneuver-level and trajectory-level uncertainty at merges and intersections.
\textit{Context-Aware Navigation}~\cite{robust-nav-2025} uses a behavior tree to switch planners and controllers based on weather, terrain, and sensor health---maintaining stability when fixed pipelines fail.
EcoFusion~\cite{energy-aware-2023} applies similar ideas to sensor fusion, selecting early, late, or single-sensor modes by scene context. The result: 60\% lower energy, 58\% lower latency, and 9.5\% better detection than fixed fusion.

\vspace{2mm}
\noindent\textbf{Personalized and User-Centric Planning.}
\textit{Park4U Mate}~\cite{park4u-2021} combines interior context (baby seat requiring rear access, passenger count) with exterior context (obstacle layout, trunk accessibility) via a voice assistant. The system picks closer spots for mobility-impaired users, well-lit areas at night, and slower speeds on ice. A 35-participant study confirmed users felt safer keeping their eyes on the road during voice-guided parking.

\vspace{2mm}
\noindent\textbf{Learning-Based Contextual Adaptation.}
Data-driven planners often fail in rare scenarios underrepresented in training.
\textit{CAPS}~\cite{caps-2025} clusters driving scenarios into 64 codebook IDs via VQ-VAE, then upweights rare contexts during imitation learning. On Bench2Drive (220 scenarios), driving score rises from 62.26 to 68.91 and success rate from 54.16\% to 56.97\%.
Meta-RL~\cite{meta-rl-2019} takes a different route: MAML trains an adaptive controller across towns, weather, and traffic densities, enabling few-shot adaptation that outperforms both pre-trained and from-scratch RL.

\vspace{2mm}
\noindent\textbf{Takeaway.}
Table~\ref{tab:context-planning-control} summarizes these approaches. Semantic reasoning (LLMs, layered controllers) encodes social norms; behavior trees and dynamic fusion adapt to weather and energy constraints; interior/exterior context personalizes parking; VQ-VAE and MAML improve rare-scenario generalization. Open problems remain: unified context representations, validation of subjective norms, and latency of LLM/MPC inference.

\begin{figure*}[!htp]
	\centering
	\includegraphics[width=.8\textwidth]{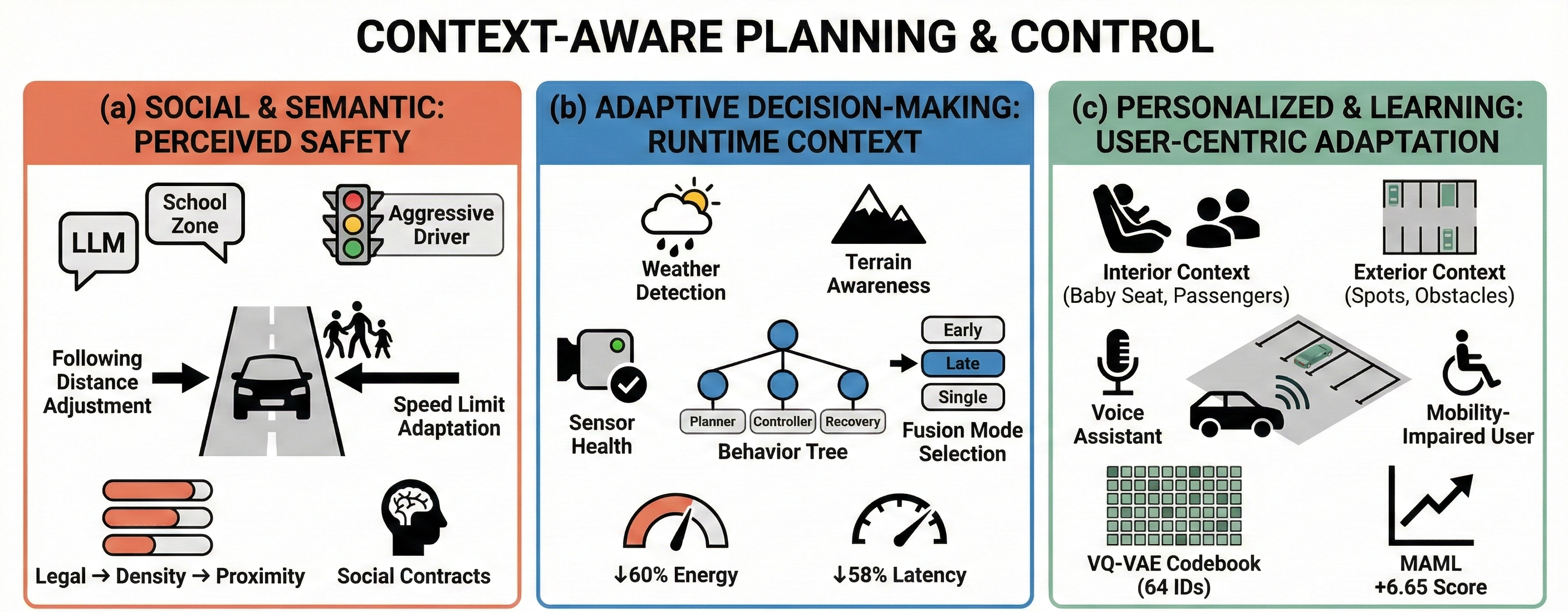}
	\caption{Context-aware Planning \& Control.}
	\label{fig:pnc}
\end{figure*}

\begin{table*}[t]
	\centering
	\small
	\caption{Representative context-aware planning \& control approaches.}
	\label{tab:context-planning-control}
	\begin{tabularx}{\linewidth}{@{}p{0.18\linewidth} p{0.26\linewidth} p{0.34\linewidth} p{0.18\linewidth}@{}}
		\toprule
		\textbf{Work}                                                                 & \textbf{Key Context Signals} & \textbf{Mechanism} & \textbf{Impact} \\
		\midrule
		\textit{Pedestrian-Aware Speed Control~\cite{ped-speed-2022}}                 &
		Pedestrian density, proximity, legal limits                                   &
		3-layer speed modulation: legal $\rightarrow$ density $\rightarrow$ proximity &
		Perceived safety via conservative behavior                                                                                                          \\
		\addlinespace[0.35em]
		\textit{LLM-Aided Social Compliance~\cite{llm-social-2025}}                   &
		High-level semantics (e.g., ``aggressive driver,'' ``school zone'')           &
		LLM reasoning $\rightarrow$ safety-compliant control policies                 &
		Human-aligned social norm compliance                                                                                                                \\
		\addlinespace[0.35em]
		\textit{Context-Aware Navigation~\cite{robust-nav-2025}}                      &
		Weather, terrain, slip, sensor status                                         &
		Behavior tree for planner/controller switching                                &
		Smoother paths; $\downarrow$ curvature                                                                                                              \\
		\addlinespace[0.35em]
		\textit{Park4U Mate~\cite{park4u-2021}}                                       &
		Interior (baby seat, passengers) + exterior (obstacles, access)               &
		Voice-based assistant with constraint optimization                            &
		Validated by a 35-user study                                                                                                                          \\
		\addlinespace[0.35em]
		\textit{CAPS~\cite{caps-2025}}                                                &
		Scenario clusters via VQ-VAE (64 codebook IDs)                                &
		Context-aware prioritization in imitation learning                            &
		+6.65 driving score; +2.81\% success rate                                                                                                           \\
		\addlinespace[0.35em]
		\textit{Predictive Motion Planning~\cite{predictive-planning-2024}}           &
		Multi-modal predictions, trajectory uncertainties                             &
		IAIMM-KF prediction + MPC optimization                                        &
		Handles maneuver \& trajectory uncertainty                                                                                                          \\
		\addlinespace[0.35em]
		\textit{Meta-RL Adaptation~\cite{meta-rl-2019}}                               &
		Task complexity, traffic density, weather                                     &
		MAML + adaptive neural controller                                             &
		Faster adaptation than baseline RL                                                                                                                  \\
		\addlinespace[0.35em]
		\textit{Energy-Efficient Systems~\cite{energy-aware-2023}}                    &
		Scene context for sensor/fusion selection                                     &
		Dynamic early/late/single-sensor fusion                                       &
		$\sim$60\% $\downarrow$ energy; +9.5\% detection                                                                                                    \\
		\bottomrule
	\end{tabularx}
\end{table*}

\section{Context-Aware Safety and Security}
\label{sec:safety}

By designing vehicle systems to be context-aware, engineers can improve hazard detection, risk assessment, anomaly detection, and overall system resilience. This section provides a broad overview of how context is being integrated into vehicle safety and security mechanisms, highlighting recent research advances (summarized in Table \ref{tab:context-assessment}).

\begin{table*}[t]
\centering
\small
\caption{Representative context-aware assessment \& risk monitoring.}
\label{tab:context-assessment}
\begin{tabularx}{\linewidth}{@{}p{0.18\linewidth} p{0.26\linewidth} p{0.34\linewidth} p{0.18\linewidth}@{}}
\toprule
\textbf{Work} & \textbf{Key Context Signals} & \textbf{Mechanism} & \textbf{Impact} \\
\midrule
\textit{INSIGHT~\cite{chen2025insight}} &
Visual data with semantic context (textual scene descriptors) &
VLM-based hazard detection &
$\uparrow$ edge-case recall/hazard AP \\
\addlinespace[0.35em]
\textit{Context-Aware Risk Indicator~\cite{tian2025context}} &
Weather, traffic, scene risk factors &
Context-Aware Risk Index computation &
$\downarrow$ collisions; $\uparrow$ comfort / efficiency \\
\addlinespace[0.35em]
\textit{CADD~\cite{chen2024context}} &
Road inclination, tire slippage, total mass &
Cross-validated context estimation \& matching &
$>\!96\%$ recall, $<\!0.5\%$ FPR; pinpoints anomalous data groups \\
\bottomrule
\end{tabularx}
\end{table*}

\vspace{2mm}
\noindent\textbf{Context for Hazard Detection and Risk Mitigation.} 
Context-aware systems significantly enhance safety in intelligent vehicles by improving perception and decision-making~\cite{zaboli2023survey}. The INSIGHT framework~\cite{chen2025insight} integrates vision and language data to interpret edge-case hazards like unpredictable pedestrians, using hierarchical attention to improve detection accuracy on rare scenarios. Similarly, a Context-aware Risk Index (CRI)~\cite{tian2025context} computes real-time risk from contextual factors—such as traffic, weather, and object motion—enabling dynamic modulation of driving behavior. Experiments in the CARLA simulator showed CRI reduced collisions and improved efficiency.

In off-road or unstructured domains, context helps discriminate traversable terrain from actual obstacles. A method for collision-aware traversability~\cite{philippe2025collision} uses a deformability-aware cost map derived from LiDAR and multispectral data to assess dynamic farm environments, reducing unsafe routing decisions. Additionally, the environmental context supports perception resilience. A system that dynamically reconfigures sensor fusion weights based on environmental noise~\cite{alikhani2024work} (e.g., prioritizing using LiDAR during heavy rain) can mitigate the impact of sensor-specific degradation.

\vspace{2mm}
\noindent\textbf{Context for Anomaly Detection and System Security.}
Beyond safety, context serves as a ``ground truth" for cyber-physical security. Context-aware anomaly detection systems cross-validate incoming sensor data against physical and environmental expectations to detect faults or attacks~\cite{chen2022context, chen2024context}. For example, a radar reporting a sudden object not seen by other sensors may signal a spoofed reading. When confirmed, context-informed controllers smoothly transition the system into a safe mode.

Intrusion detection also benefits from contextual awareness. RAIDS~\cite{jiang2019road} uses road type and driving intent to validate in-vehicle network messages; a command for sharp steering while on a high-speed highway is flagged as an anomaly. Additionally, context allows for \textit{semantic consistency checking}. If a vehicle detects a stop-sign via computer vision but the V2X data from the ITS indicates no such intersection exists~\cite{chen2022context}, the system can trigger a verification routine, such as to detect potential ``Adversarial Patch" attacks on the camera~\cite{xiang2021patchguard}.

\vspace{2mm}
\noindent\textbf{Dynamic Safety Policy and Remediation.}
Context also governs when autonomy should yield to human drivers. Competence-aware systems~\cite{mahmud2023semi} monitor for low-confidence scenarios (e.g., novel construction or weather) and initiate handovers, improving safety and driver trust. Overall, these examples show that integrating scene semantics, environment status, and internal system confidence into vehicle decision loops enhances both operational robustness and hazard anticipation.

Finally, context enables adaptive safety envelopes. A context-aware monitor~\cite{haupt2021towards} can proactively tighten braking thresholds or expand the ``keep-out" zone around the vehicle during high-risk contexts, such as driving near a school zone or on black ice. When an inconsistency is detected, context-informed controllers do not simply halt the vehicle; they transition the system into a context-appropriate safe state (e.g., pulling over to a wide shoulder rather than stopping in a narrow lane), thereby balancing security with operational availability.

\section{Context-Aware Connected Vehicles}
\label{sec:v2x}

With the rapid development and widespread adoption of V2X communication~\cite{zhou2020evolutionary}, as well as the growing intelligence of vehicles and infrastructure~\cite{shi2024soar, cress2023intelligent}, the acquisition and use of contextual information have become increasingly comprehensive and diverse. Connected vehicles achieve context awareness through an end-to-end pipeline that defines how information is acquired, communicated, fused, and ultimately used by driving functions. Across this pipeline, research has progressed from broadcasting ego-state and basic hazard cues to sharing rich collective perception and predictive intent, and finally to closing the loop by feeding fused context into planning and control (summarized in Table~\ref{tab:context_aware_v2x}).

\subsection{Acquisition of context information}

The acquisition of context information by connected vehicles has evolved from periodic sharing of ego status and event cues to collaborative sensing of the environment. Early systems mostly sent updates on their own status or simple alerts about events, which helped drivers and automated systems avoid conflicts and coordinate on the road~\cite{huang2011implementation, fallah2012cyber}. Subsequent studies showed that infrastructure and nearby vehicles can contribute environment-level information, such as objects~\cite{ zhang2023robust, chen2019f}, free space~\cite{zhang2021emp, he2021vi, ren2025unisense}, drivable topology~\cite{he2023vi, ahmad2020carmap, cui2024vilam}, and environment information~\cite{ilie2020hoa} derived from cameras, LiDAR, radar, and internet interfaces. A core insight is that “who sees what” is complementary: viewing points differ in occlusions and sensor noise, so the value of each observation depends on geometry, weather, and traffic density. Recent studies focus on sharing information only when it adds something useful or new, making communication more efficient without missing out on important details for driving decisions~\cite{rehman2022ctmf, hu2022where2comm}. Some efforts also standardize the semantics of shared entities by agreeing on standard terms for objects, road layouts, and traffic signals, and adding precise time and location data to make combining information from different sources easier and more accurate~\cite{shi2022vips, wang2023vimi}.

\subsection{Transmission of context information}

Transmission in context-aware connected vehicles is about delivering the right context to the right recipients at the right time under strict latency and bandwidth constraints. Recent work treats communication as an information-centric resource, scheduling and shaping messages for downstream applications such as perception, prediction, and control~\cite{zhang2025vehicle, martinez2022preliminary}. Specifically, context-aware transmitters adapt rate, fidelity, and recipients according to local scene complexity, channel load, and task salience~\cite{sharma2020context}. \cite{chaudhry2023toward} only sends alerts if there is a real risk, like a possible collision. \cite{dayal2021adaptive} reduces details about distant objects that are less important. \cite{liu2020when2com} shares information only with vehicles that need it, helping avoid network congestion. Some works prioritize contents by spanning object lists and weighting information value~\cite{yang2023what2comm, su2024collaborative}. At the networking layer, cross-layer designs couple application priorities with MAC scheduling and congestion control to ensure bounded delay in safety-critical contexts~\cite{math2017v2x, lee2024causality}. Techniques such as deadline-aware retransmission~\cite{fouda2024harq}, predictive prefetching using short-term intent~\cite{wang2025cmp}, and multipath/hybrid links~\cite{yacheur2023drl} mitigate variability and packet loss.  

\subsection{Fusion of context information}
Context-aware fusion in connected vehicles treats all incoming messages (e.g., onboard detections, cross-vehicle observations, and map priors) as information whose value depends on scene semantics, timing, and downstream task needs. Some works~\cite{taddei2024multi,eiter2019towards} introduce a common representation in the form of a layered local dynamic map, which integrates static geometry, traffic rules, and dynamic entities. This structured representation helps reduce data association errors and improves robustness under occlusion. Object-level fusion aligns tracks using filters that consider time delay, pose noise, and heterogeneous sensors to improve multi-object tracking across agents~\cite{chiu2024probabilistic, shi2022vips}. Feature-level fusion projects intermediate neural features into a shared bird-eye view and select task-relevant cues for processing~\cite{xu2022cobevt, xu2022v2x}. Some works propose utility-aware gating~\cite{hu2022where2comm, hu2024communication} that prioritize information based on its value to the task. These policies emphasize occluded hazards, conflict zones, and regions of high uncertainty to achieve bandwidth savings with minimal impact on downstream application performance. In fleets and infrastructure scenarios, where agents contribute heterogeneous information, domain-invariant representations~\cite{li2024di} are used to mitigate distribution shifts between vehicles and infrastructure. Finally, plausibility checks~\cite{liu2021miso} in kinematic feasibility and map consistency downweight contradictory reports before they contaminate the shared scene, improving the resilience to faults, latency outliers, and adversarial inputs. 

\subsection{Utilization of context information}
Context-aware utilization turns fused information into action by weighing it against meaning, uncertainty, and task goals. Some research~\cite{zhang2023occlusion, muller2022motion} raises risk levels in areas drivers cannot see and gives less trust in unclear signals at blocked intersections during path planning. Other studies~\cite{wang2020v2vnet, wang2025cmp} weight input from others based on their relevance and reliability. This helps clarify where vehicles are going, making it easier to brake safely and judge traffic gaps. In car-following scenarios, controllers leverage shared state and intent to enable delay-aware cooperative adaptive cruise control~\cite{oncu2014cooperative, hua2024communication} and robust model predictive control~\cite{zhang2023robust}. These designs help maintain the following stability while improving safety, ride comfort, and energy efficiency. Some works~\cite{an2018design, luo2019cooperative} propose trajectory planners that explicitly account for V2V delay to achieve higher success rates and less disruption to surrounding traffic. In human driving scenarios, recent work shows that context-aware warnings about vulnerable road users can encourage timely yielding and reduce crash risk without overwhelming attention~\cite{angulo2021evaluation}. Beyond driving control, context information is also used in connected vehicle applications such as navigating unsignalized intersections~\cite{vahdat2023catmi} and managing electric vehicle charging~\cite{iqbal2025context}.

\begin{table*}[t]
\centering
\small
\caption{Summary of context-aware connected vehicle approaches.}
\label{tab:context_aware_v2x}
\begin{tabularx}{\textwidth}{@{}l p{2.2cm} p{2.9cm} X@{}}
\toprule
\textbf{Pipeline Stage} & \textbf{Work} & \textbf{Focus Area} & \textbf{Key Concepts \& Mechanisms} \\ \midrule
\multirow{6}{*}{\textit{Acquisition}} & \cite{huang2011implementation, fallah2012cyber, zhang2023robust, chen2019f, zhang2021emp, he2021vi, ren2025unisense, he2023vi, ahmad2020carmap, cui2024vilam, ilie2020hoa} & Collective Perception & Shifting from ego-state/hazard alerts to sharing rich environment data (objects, free space, and topology) from multi-modal sensors. \\ \cmidrule(l){2-4} 
 & \cite{rehman2022ctmf, hu2022where2comm, shi2022vips, wang2023vimi} & Efficient Sensing & Adaptive information sharing based on utility; semantic standardization and precise spatio-temporal alignment. \\ \midrule
\multirow{6}{*}{\textit{Transmission}} & \cite{zhang2025vehicle, martinez2022preliminary, sharma2020context, chaudhry2023toward, dayal2021adaptive, liu2020when2com, yang2023what2comm, su2024collaborative} & Task-Oriented Messaging & Information-centric resource scheduling; adapting rate, fidelity, and recipients based on task salience and scene complexity. \\ \cmidrule(l){2-4} 
 & \cite{math2017v2x, lee2024causality, fouda2024harq, wang2025cmp, yacheur2023drl} & Reliable Networking & Cross-layer design for bounded delay; predictive prefetching and deadline-aware retransmission to mitigate packet loss. \\ \midrule
\multirow{5}{*}{\textit{Fusion}} & \cite{taddei2024multi, eiter2019towards, chiu2024probabilistic, shi2022vips, xu2022cobevt, xu2022v2x, li2024di} & Multi-modal Integration & Layered local dynamic maps; BEV feature-level fusion; domain-invariant representations to mitigate distribution shifts. \\ \cmidrule(l){2-4} 
 & \cite{hu2022where2comm, hu2024communication, liu2021miso} & Quality \& Resilience & Utility-aware gating and plausibility checks to ensure resilience against latency outliers and adversarial inputs. \\ \midrule
\multirow{6}{*}{\textit{Utilization}} & \cite{zhang2023occlusion, muller2022motion, wang2020v2vnet, wang2025cmp, angulo2021evaluation} & Risk-Aware Planning & Weighting external information by reliability; accounting for occlusions and human factors in decision making. \\ \cmidrule(l){2-4} 
 & \cite{oncu2014cooperative, hua2024communication, zhang2023robust, an2018design, luo2019cooperative, vahdat2023catmi, iqbal2025context} & Cooperative Control & Delay-aware CACC and robust MPC; trajectory planning for intersections and fleet management under V2X constraints. \\ \bottomrule
\end{tabularx}
\end{table*}


\section{Trend and Evolvement of Context}
\label{sec:trend}


Modern intelligent vehicle computing stacks are shifting from isolated, frame-by-frame perception and fixed heuristics to context-rich, temporally grounded, and risk-aware intelligence. By injecting context—spanning temporal dynamics, operational design domain (ODD), physical consistency, uncertainty, intent, and user preferences, systems can both \emph{improve reliability} and \emph{lower end-to-end latency/bandwidth}. Context enables robustness (cross-sensor/scene validation), efficiency (ROI-aware anytime inference), and enhance safety (uncertainty- and risk-driven behavior), while ultimately delivering smoother, more personalized driving experiences.

\begin{table*}[t]
\centering
\small
\caption{Trend of ``Context'' in Intelligent Vehicles}
\label{tab:context-trends}
\begin{tabular}{p{0.12\linewidth} p{0.25\linewidth} p{0.33\linewidth} p{0.22\linewidth}}
\toprule
\textbf{Trend} & \textbf{What changes} & \textbf{Typical signals/tech} & \textbf{Benefits} \\
\midrule
Rules Learned & Heuristics $\rightarrow$ embeddings & Rule-based, features & Fewer handcrafts, faster iteration \\
\addlinespace[0.4em]
3D $\rightarrow$ 4D & Frames $\rightarrow$ state over time & BEV temporal fusion, world models, occupancy flow & Stable tracks/intents in clutter \\
\addlinespace[0.4em]
Efficiency & Uniform $\rightarrow$ context-aware & ROI inference, anytime DNNs & Lower latency \& bandwidth \\
\addlinespace[0.4em]
Safety & Static thresholds $\rightarrow$ ODD-aware & Uncertainty, risk maps & Fewer interventions, clearer rationale \\
\addlinespace[0.4em]
Robust/Secure & Single-view $\rightarrow$ cross-validation & Physics/scene consistency, anomaly detection & Fault/spoof resilience \\
\addlinespace[0.4em]
Application & Generic $\rightarrow$ personalized & Voice queries, rules-aware UX & Smoother, convenient driving \\
\bottomrule
\end{tabular}
\end{table*}

Across the autonomy stack, \emph{context} is increasingly treated as a first-class signal that reshapes how intelligent vehicles learn, perceive, decide, and interact:

\vspace{2mm}
\noindent\textbf{(1) Rules $\rightarrow$ Learned representations.}
Hand-tuned heuristics are being replaced by learned embeddings: instead of fixed gap-acceptance or lane-change thresholds, policies consume scene/agent embeddings (often self-supervised and map-conditioned), reducing manual feature engineering and accelerating iteration.

\vspace{2mm}
\noindent\textbf{(2) 3D frames $\rightarrow$ 4D state over time.}
Perception and planning are shifting from per-frame 3D outputs to a temporally grounded world state. Temporal BEV fusion, occupancy flow, and world models stabilize tracks through clutter and occlusion and better capture intent, enabling smoother long-horizon planning.

\vspace{2mm}
\noindent\textbf{(3) Uniform efficiency $\rightarrow$ Context-aware efficiency.}
Compute and sensing budgets are becoming adaptive. ROI inference and anytime DNNs allocate more resolution/compute to critical regions while pruning tokens/tiles elsewhere, cutting latency and bandwidth without degrading task quality.

\vspace{2mm}
\noindent\textbf{(4) Static thresholds $\rightarrow$ ODD-aware safety.}
Safety logic is moving beyond fixed cutoffs to operating-condition-aware control. Uncertainty estimates and risk maps inform adaptive behaviors (e.g., larger headways, cautious creeping) and provide clearer rationale for interventions and fallbacks.

\vspace{2mm}
\noindent\textbf{(5) Single-view $\rightarrow$ Cross-validated robustness and security.}
Reliability increasingly comes from redundancy and consistency checks rather than any single sensor/view. Physics/scene consistency, cross-sensor agreement, and anomaly detection improve resilience to faults, distribution shift, and spoofed cues.

\vspace{2mm}
\noindent\textbf{(6) Generic $\rightarrow$ Personalized, rules-aware applications.}
User-facing autonomy is becoming more personalized via voice/intent interfaces and preference conditioning, while remaining compliant with traffic rules and safety constraints—supporting smoother, more convenient driving.

\vspace{2mm}
\noindent\textbf{Summary.}
Injecting context—learned representations, temporal state, adaptive efficiency, ODD-aware risk, cross-validation, and user intent—systematically improves capability and reliability while keeping autonomy efficient and bounded in real time.

\section{Challenges and Opportunities}
\label{sec:challenge}

Modern intelligent vehicles increasingly rely on context-aware designs to interpret complex driving scenarios. A context-aware intelligent driving system continuously integrates information from perception, prediction, planning, control, and safety/security modules to form a coherent picture of “what is happening” and “what is likely to happen next.” 

In this section, we discuss key challenges in achieving such context awareness across all major aspects of an intelligent driving pipeline, as well as the potential opportunities these challenges unlock for advancing vehicle intelligence. We focus on research challenges – multimodal context fusion, temporal context modeling, rare-event handling under resource constraints, and collaborative context sharing – highlighting both technical difficulties and promising directions for each.

\subsection{Multi-Modal Context Fusion}

Perception is the foundation of context-aware driving, as it converts raw sensor data into an understanding of the environment. A central challenge in context-aware perception is multimodal context fusion – combining heterogeneous sensor inputs and contextual data into a unified “context engine.” Intelligent vehicles are equipped with diverse sensors that produce very different data representations: time-series signals from CAN bus, GPS, or IMU; dense image matrices from cameras; sparse point clouds from LiDAR and radar; textual map information, etc. Each modality has unique strengths and limitations – for example, LiDAR excels at precise distance measurement while cameras provide rich semantic cues, and CAN/IMU sensors report the vehicle’s internal states. These data streams also vary in granularity and reliability (e.g., cameras and LiDAR can be affected by lighting or weather conditions). 

Fusing such \textit{heterogeneous inputs} into a consistent context is non-trivial. Designing effective fusion strategies remains an open problem: simple rule-based integration may fail to capture complex cross-modal correlations, whereas learning-based fusion (using dedicated encoders for each modality) requires large datasets and careful calibration. A hybrid approach that leverages both domain knowledge and data-driven learning is often considered, but finding the right balance is challenging. Despite these difficulties, progress in multimodal fusion promises significant opportunities. Robust context-aware perception can yield more reliable environment models than any single sensor alone – for instance, combining camera semantics with LiDAR depth can improve object detection and classification, and incorporating HD maps or traffic signal data can provide prior context for interpreting sensor readings. In effect, advanced fusion can enable the vehicle to “see” the environment with greater clarity and resilience (even under adverse conditions), forming a strong basis for downstream prediction and planning.

A promising direction is an adaptive anytime fusion system\cite{dean20201,zilberstein1996using,shazeer2017outrageously} that exposes multiple fusion points and selects among them online along two design dimensions: \emph{scope} (intra-vehicle versus inter-vehicle/infrastructure fusion) and \emph{stage} (early/raw, mid/feature, late/semantic fusion). Concretely, the stack can maintain parallel fusion paths at different stages within the vehicle, and opportunistically incorporate V2X-shared context when communication is available and trusted. The fusion policy then adapts to accuracy/latency requirements and contextual cues such as ODD complexity, traffic density, visibility, weather, modality health/uncertainty, and compute/bandwidth budgets. Under tight real-time constraints, the system may prioritize lightweight mid- or late-stage intra-vehicle fusion; when risk increases (e.g., occlusions, degraded visibility, complex interactions), it can escalate to richer early-stage fusion and broaden the scope to inter-vehicle/infrastructure context to extend perception beyond line-of-sight. This context-driven, multi-path design provides graceful degradation under missing modalities while enabling principled accuracy--latency trade-offs.


\subsection{Temporal Context Modeling}

Building on perception, prediction modules use context to forecast the future state of the environment – for example, predicting the trajectories and intentions of other road users. A key challenge here is capturing temporal context: the system must maintain an ever-updating understanding of “what’s happening now and what’s likely next” over various time scales. This involves modeling how the scene evolves, rather than just analyzing static snapshots. 

Technically, one needs to represent context at \textit{multiple} temporal horizons (from instantaneous events up to long-range trends on the order of tens of seconds). Recent approaches emphasize constructing a \textit{4D world memory} that accumulates spatio-temporal information – for instance, a bird’s-eye-view occupancy flow that logs moving objects over time, combined with the ego vehicle’s state, map data (road layouts, traffic rules), and even weather/visibility conditions. Another promising representation is the \textit{dynamic scene graph}, a structured temporal graph where nodes correspond to relevant entities (lanes, vehicles, pedestrians, traffic signals, etc.) and edges capture relationships like yielding, occlusion, or right-of-way interactions. Crucially, such representations must handle uncertainty and causal relationships in the evolving scene. 

Context-aware prediction also extends to understanding internal driver/passenger states over time – for example, tracking a human driver’s gaze or fatigue levels in relation to external events (the AIDE project demonstrates value in fusing interior and exterior context). Effectively leveraging temporal context yields clear opportunities: it enables the intelligent vehicle system to anticipate and predict rare or subtle events that purely reactive systems might miss. For instance, by analyzing temporal sequences, a context-aware predictor can foresee a pedestrian’s intention to jaywalk or detect a developing hazard (near-miss scenario) a few seconds in advance. Many challenging driving maneuvers benefit from rich temporal context, including merging into fast traffic, handling an unprotected left turn, reasoning about occluded vehicles, or planning a long-horizon lane change. By incorporating multi-second context, the vehicle can better infer other agents’ intentions and make more human-like, foresighted predictions (e.g,. anticipating that a car in an adjacent lane might suddenly cut in). 

In summary, while temporal modeling increases system complexity (requiring state memory and sequence learning under uncertainty), it offers the opportunity for significantly improved prediction accuracy and safer, smoother interactions on the road.

\subsection{Rare-Event Handling}

Given rich contextual information from perception and prediction, the planning and control module of an intelligent vehicle must decide how to act – selecting safe and efficient maneuvers in response to the current and predicted context. Context-aware planning faces the challenge of reasoning about myriad contextual factors in real time. The vehicle not only needs to follow traffic rules and navigate to its destination, but it must also adapt to dynamic agent behaviors, road context (e.g, upcoming construction or road topology), and even its own performance limits (e.g, degraded sensor functionality). One major research problem is handling rare critical events versus routine driving under strict computational and operational constraints. Intelligent vehicles can operate continuously (24/7), but truly critical or abnormal situations (e.g, sudden pedestrian crossings, tire blowouts, or other emergencies) occur infrequently. 

It is infeasible to run every advanced perception-prediction algorithm at full capacity at all times due to limits in onboard computing power, energy, and thermal dissipation. Thus, a promising strategy is context-aware resource management: the system can sense and think on demand. In normal conditions, the planning/control pipeline might use an “essential sensor set” and cached inferences, running in an energy-efficient mode. When the context engine detects a potentially critical situation or an anomaly, it can trigger heightened awareness – activating additional sensors or high-fidelity models (“adaptive event-driven sensing”) and even speculative inference to proactively evaluate possible outcomes. This dynamic allocation of computational effort is an opportunity to reconcile safety with efficiency: the vehicle conserves resources during mundane driving, yet remains ready to ramp up its analytical depth when the context signals elevated risk. 

Another challenge in context-aware planning is ensuring the system can degrade gracefully when context changes or information is lost. For example, if certain sensors fail or a communication link drops, the planning module should still function (perhaps with reduced capabilities) instead of catastrophically breaking. Context-awareness can aid here by recognizing changes in available context and adjusting the control strategy accordingly (e.g., slowing down and adopting a conservative stance if critical context information becomes uncertain). 

Overall, infusing planning and control with context enables more anticipatory and adaptive decision-making. A context-rich planner can make smoother maneuvers – for instance, starting to slow down earlier if it predicts a traffic light will turn red based on timing and map context, or preemptively choosing a safer lane in anticipation of an upcoming merge. It also opens opportunities for higher-level optimization, such as balancing comfort and efficiency based on situational context (driving more cautiously in a school zone or in bad weather, but reverting to normal behavior otherwise). The technical difficulties lie in designing planners that can digest complex context inputs and respond in real time, but success in this area would greatly improve both the safety and the naturalness of intelligent vehicle behavior.

\subsection{Collaborative Context Sharing}

Beyond individual vehicles, a significant opportunity for enhancing safety lies in collaborative context sharing among vehicles (and with infrastructure). Intelligent vehicles can communicate to jointly build and maintain a shared 4D context of the road scene. 

The vision is that a fleet of nearby cars, along with roadside sensors, could exchange information to create a collective view covering a wider area and range of visibility than any single vehicle’s perspective. Each vehicle could benefit from this aggregate context – for instance, receiving warnings about an unseen hazard or the movements of a distant aggressive driver, thus improving its own safety and efficiency. However, realizing this opportunity comes with considerable challenges. One issue is defining a common language or abstraction for shared context. Vehicles may have different sensor suites and internal models; they need standardized ways to represent and communicate context (objects, events, uncertainties) so that information from one can meaningfully augment another’s understanding. Another challenge is the management of information exchange under practical constraints – communication bandwidth and latency are limited, connections may be intermittent, and not all contexts are equally critical. Effective protocols must prioritize the most relevant and safety-critical information, ensure it is fresh (expiring outdated data), and handle synchronization so that the shared situational picture remains consistent and causal despite network delays or jitter. 

Security is also a crucial facet of collaborative context sharing. Relying on external information introduces risks: false or malicious data could lead to wrong decisions. Therefore, context-sharing frameworks must incorporate trust, provenance tracking, and conflict resolution at scale. For example, vehicles should be able to verify the source of incoming context data and resolve discrepancies if two sources report conflicting situations. Research in this area overlaps with cybersecurity and distributed consensus, aiming to make sure that shared context improves safety without opening new vulnerabilities. If these technical hurdles can be overcome, the payoff would be transformative – a network of context-aware vehicles acting cooperatively could prevent accidents by effectively “seeing around corners” and learning from each other’s experiences in real time. This level of collective intelligence and foresight represents a major opportunity to push intelligent vehicle safety beyond the capability of any standalone system. 

In summary, context-aware design is an emerging paradigm that spans all components of the intelligent vehicle software stack, introducing challenges at every level from sensing to decision-making. Tackling issues like multimodal fusion, temporal modeling, resource-aware operation, and collaborative context integration is difficult, but it is also key to closing the gap between current intelligent vehicles and human-level driving performance. By persistently addressing these challenges, researchers can unlock opportunities for more robust perception, more predictive behavior modeling, smarter planning under uncertainty, and safer, coordinated driving – ultimately advancing intelligent vehicles toward greater reliability and intelligence on our roads.

\section{Conclusion}
\label{sec:conclusion}

Intelligent vehicles are increasingly capable of supporting adaptive, flexible applications beyond driving themselves. Emerging personalized and context-aware functionalities—such as context-aware ADAS, in-cabin monitoring, and fleet management—depend on rich contextual information. In this paper, we systematically review state-of-the-art (SOTA) context-aware methods spanning (i) environment understanding, (ii) planning and control, (iii) safety and security, and (iv) connected vehicles. Based on a thorough analysis of recent trends in context-aware design, we identify four key technical challenges in building a contextual engine for future intelligent vehicles: multimodal context fusion, temporal context modeling, rare-event handling, and collaborative context sharing. We hope this survey will motivate further research and development of context-aware applications for next-generation intelligent vehicles.

\bibliographystyle{IEEEtran}
\bibliography{main}

\end{document}